\documentclass{article}
\usepackage[T1]{fontenc}
\usepackage[preprint]{neurips_2026}
\usepackage{graphicx}
\usepackage{booktabs}
\usepackage{tabularx}
\usepackage{array}
\usepackage{adjustbox}
\usepackage{pdflscape}
\usepackage{hyperref}
\usepackage{url}
\hypersetup{hidelinks}

\makeatletter
\renewcommand{\@noticestring}{}
\makeatother

\title{BenchBench-Protocol:\\Evaluating \mbox{Real-World} \mbox{Wet-Lab} Protocol\\Reasoning and Modification}

\author{
Aditya Sivakumar\textsuperscript{*} \\
Benchling \\
\And
Ashu Singhal \\
Benchling \\
\AND
Nicholas Larus-Stone \\
Benchling \\
\And
Nithin Parsan\textsuperscript{*,\textdagger} \\
Benchling
}

\newcommand{\scoreatk}{score@$k$}
\newcommand{\bestatk}{best-of-$k$}
\newcolumntype{Y}{>{\raggedright\arraybackslash}X}

\begin{document}
\maketitle
\begingroup
\renewcommand{\thefootnote}{\fnsymbol{footnote}}
\footnotetext[1]{Equal contribution.}
\footnotetext[2]{Corresponding author and team lead: \texttt{nithin.parsan@benchling.com}.}
\endgroup

\begin{abstract}
We introduce BenchBench-Protocol, a benchmark for large language models of 149 protocol-modification tasks recovered from modifications that scientists made to published protocols during real experimental work. Adapting a published protocol to a new experiment is a routine task for a wet-lab scientist, and a correct modification requires accounting for prior choices and downstream steps. Recent life-science benchmarks have moved toward open-ended, rubric-graded tasks, but tasks are typically elicited from experts rather than reconstructed from real-world modifications. BenchBench-Protocol tasks are derived from differences between a published protocol and a version a scientist modified, which provides the basis for the query and the weighted rubric elements for a correct response. The benchmark draws from 96 source protocols across nine domains of wet-lab biology and only includes tasks rated highly after review by domain experts. We evaluate nine closed and open models; Claude Opus 5 scores highest at 59.2\% normalized rubric score, with other models between 34.1\% and 47.1\%, and the benchmark remains unsaturated when taking the best of ten attempts. As models are increasingly helpful in life-sciences research, evaluating them on routine wet-lab tasks becomes correspondingly important. We present BenchBench-Protocol as both a grounded assessment of wet-lab reasoning and evidence for the utility of real-world experiments to construct benchmark tasks.
\end{abstract}

\section{Introduction}

Adapting an existing laboratory protocol for a new experiment is a common task for wet-lab scientists. Correct decisions require reasoning over previous choices and intended next steps, not just recalling facts or retrieving existing knowledge. Therefore, an important practical skill is being able to modify or extend an existing protocol while anticipating its downstream effects.

Recent frontier benchmarks of language models for life-science research have moved toward more open-ended, rubric-graded tasks that resemble research requests. These tasks are often constructed by asking experts to write tasks representative of their work. While this process results in high-quality tasks that are scientifically correct and plausible, they may diverge from the distribution of modifications used in real-world workflows, which are constrained by available reagents, instrument access, and prior results.

\begin{figure}[t]
\centering
\includegraphics[width=\linewidth]{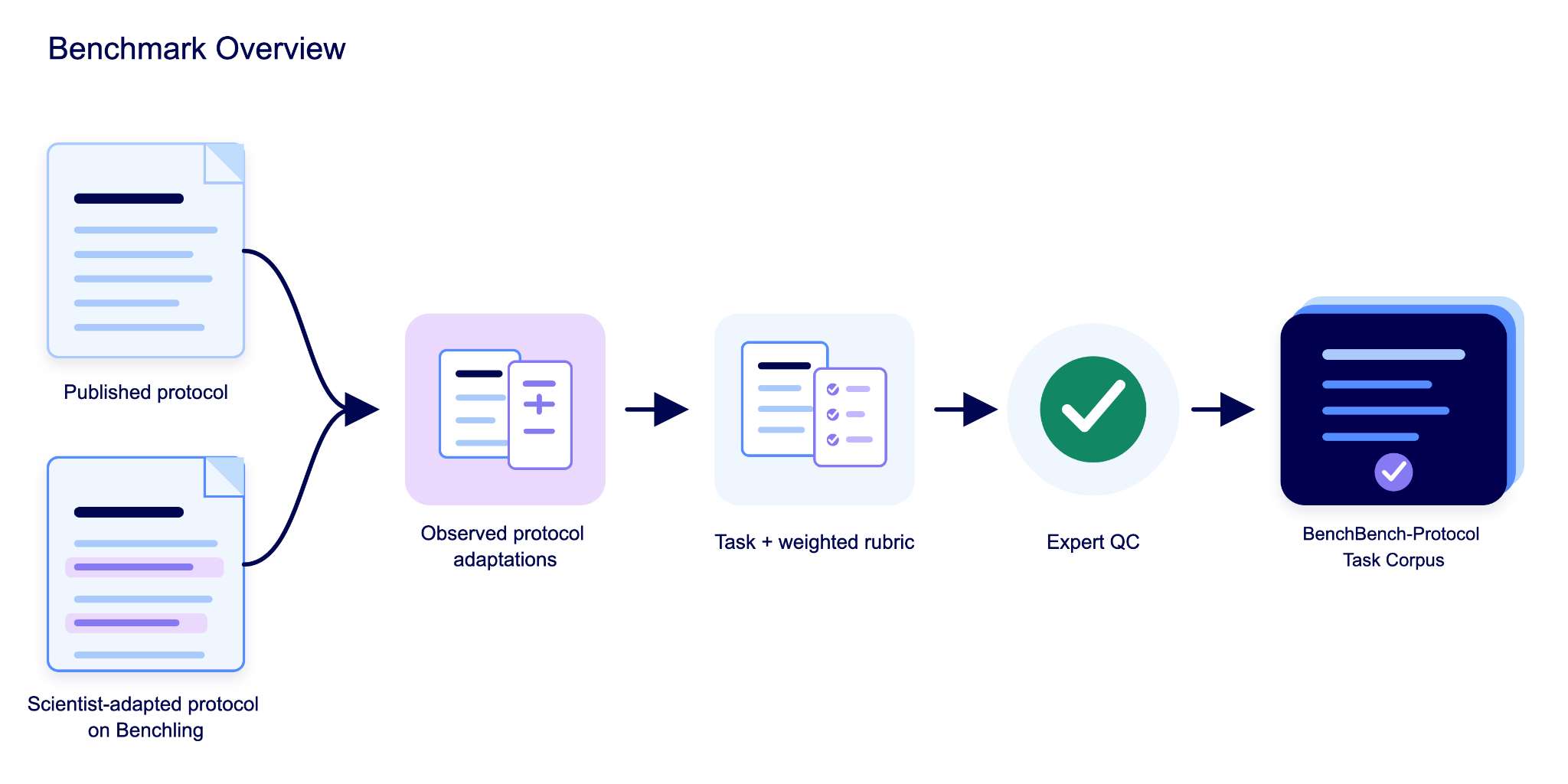}
\caption{Benchmark construction overview. Published protocols and scientist-adapted versions on Benchling from contributors are compared to produce tasks and weighted rubrics, which are retained after expert quality control.}
\label{fig:construction}
\end{figure}

We introduce BenchBench-Protocol, a benchmark of protocol-modification tasks. Rather than creating tasks from scratch, we recover them from real modifications scientists have made to published protocols. Tasks are realistic, open-ended, and free-response. Each task and rubric was then independently reviewed by multiple scientists with a PhD in a relevant field and at least three years of bench experience. Figure~\ref{fig:construction} summarizes this construction process.

We contribute:
\begin{itemize}
    \item \textbf{BenchBench-Protocol:} a benchmark of 149 expert-reviewed protocol-modification tasks spanning nine domains of wet-lab biology, phrased as a scientist would pose the question to an LLM rather than as a formal problem statement.
    \item \textbf{Weighted rubrics:} expert-verified rubrics grounded in the modifications scientists made, used to score open-ended responses against the specific downstream dependencies a correct answer must address.
\end{itemize}

\section{Related Work}

Recent benchmarks have expanded beyond simple biological fact recall and increasingly focus on the diverse tasks a scientist may perform. Several benchmarks assess the research-assistant capabilities of a model: LAB-Bench has an extensive suite of biology research tasks~\citep{laurent2024labbench}, with LABBench2 expanding these tests with free-response questions (ProtocolQA2, LitQA3, FigQA2, and others) that show whether models can troubleshoot protocols and understand and evaluate figures, tables, and data from research papers and databases~\citep{laurent2026labbench2}. BioProBench assesses model capabilities in error correction and protocol generation~\citep{liu2025bioprobench}, while TroubleshootingBench evaluates the ability to identify and correct wet-lab error scenarios created by experts~\citep{openai2025gptoss}. BioLP-bench, like ProtocolQA and TroubleshootingBench, introduces a critical mistake into a protocol and measures whether a model can identify the responsible step~\citep{ivanov2024biolpbench}.

Several benchmarks also address agentic and long-horizon capabilities for end-to-end research tasks. BixBench tests agents' ability to execute code and reason over bioinformatics results~\citep{mitchener2025bixbench}. ScienceAgentBench and AstaBench extend agentic evaluations across scientific workflows more broadly~\citep{chen2025scienceagentbench,bragg2025astabench}. GeneBench and GeneBench-Pro are frontier evaluations assessing multi-stage quantitative biology and carefully construct tasks to constrain the number of valid solution paths and permit exact verification~\citep{li2026genebench,li2026genebenchpro}. While these benchmark tasks require strong biological reasoning, most are limited to computational biology.

LifeSciBench is the closest analogous benchmark, containing expert-authored tasks with a rubric and a single-turn free-response format~\citep{liu2026lifescibench}. One of its seven workflows explicitly covers the design and optimization of assays, constructs, and protocols. LifeSciBench is constructed by subject-matter experts with strong quality audits, but, similar to LABBench2 ProtocolQA2, the plausible modifications an expert may suggest to protocols may differ from the distribution of modifications in real-world workflows. BenchBench-Protocol complements and extends the suite of life-sciences capability benchmarks by deriving tasks from modifications scientists made to published protocols during real experimentation.

\section{Benchmark Design and Construction}
\label{sec:benchmark-design}

\subsection{Task Formulation}
\label{sec:task-formulation}

In each task, the model receives a first-person, naturally phrased scientist query and a published protocol to extend or modify, and returns a free-text answer with a rationale. During evaluation, model tool use is unrestricted within the assigned search configuration described in Section~\ref{sec:evaluation-method} and Appendix~\ref{app:model-config}. Responses are graded by an LLM-as-a-judge that receives the system prompt, task, rubric, and response under evaluation. All evaluation attempts were graded with GPT-5.6 Terra at medium reasoning to standardize grading and minimize variation between model families. No significant scoring differences were observed across two representative grader models during testing (Appendix~\ref{app:grader-agreement}). Each rubric criterion is presented to the grader separately so that its score reflects whether the response specifically satisfies that criterion.

\subsection{Benchmark Coverage}

BenchBench-Protocol contains 149 evaluation tasks drawn from 96 published protocols across nine domains of wet-lab biology, listed in Table~\ref{tab:coverage}. The published protocols are compared with protocols created by contributors on Benchling, a life-sciences platform used for wet-lab work across the major life-science domains.

\begin{table}[t]
\caption{Benchmark coverage by protocol domain.\protect\\Task and source protocol counts are shown for each of the nine domains.}
\label{tab:coverage}
\centering
\small
\begin{tabularx}{\linewidth}{Yrrr}
\toprule
\textbf{Protocol domain} & \textbf{Tasks} & \textbf{Share} & \textbf{Protocols} \\
\midrule
Protein biochemistry and purification & 32 & 21.5\% & 21 \\
Cell culture and cell-based assays & 27 & 18.1\% & 16 \\
Molecular biology and cloning & 26 & 17.4\% & 17 \\
Staining, imaging, and cytometry & 24 & 16.1\% & 17 \\
Genomics, transcriptomics, and sequencing & 19 & 12.8\% & 13 \\
Microbiology and microbial methods & 14 & 9.4\% & 8 \\
Media, reagent prep, and recipe & 4 & 2.7\% & 2 \\
Model organism, in vivo, and primary tissue & 2 & 1.3\% & 1 \\
Chemical synthesis and conjugation & 1 & 0.7\% & 1 \\
\midrule
\textbf{Total} & \textbf{149} & \textbf{100.0\%} & \textbf{96} \\
\bottomrule
\end{tabularx}
\end{table}

\subsection{Benchmark Construction}

Tasks are constructed by comparing a published protocol against a version of the protocol that a scientist modified on Benchling. The differences between versions provide a record of what a scientist did to adapt the protocol for a new experimental goal. These differences supply the ground truth for the task, defining both the relevant modification and the downstream consequences that must be addressed. From each difference, a question and accompanying rubric are drafted, and a quality-control process removes questions that are unanswerable, malformed, or ambiguous. Review additionally validates that the differences are scientifically meaningful and relevant.

Tasks may carry multiple intent labels, which are not mutually exclusive, and each task is additionally assigned one primary intent. These labels allow performance to be broken down by the kind of judgment a task demands, rather than only by biological subject matter. Definitions of each source-protocol domain and task intent are provided in Appendix~\ref{app:taxonomy}. Table~\ref{tab:intents} reports primary intents.

\begin{table}[t]
\caption{Distribution of primary task intents. Task counts and shares are shown for each of the seven intents.}
\label{tab:intents}
\centering
\small
\begin{tabular}{lrr}
\toprule
\textbf{Primary task intent} & \textbf{Tasks} & \textbf{Share} \\
\midrule
Plan audit & 38 & 25.5\% \\
Measurement and result interpretation & 35 & 23.5\% \\
Troubleshooting & 29 & 19.5\% \\
Recordkeeping & 23 & 15.4\% \\
Procedure specification and adaptation & 14 & 9.4\% \\
Quantitative formulation & 6 & 4.0\% \\
Follow-up experiment & 4 & 2.7\% \\
\midrule
\textbf{Total} & \textbf{149} & \textbf{100.0\%} \\
\bottomrule
\end{tabular}
\end{table}

\subsection{Expert Review}

An initial cohort of tasks was reviewed by domain experts and refined or corrected to yield a final set of 149 questions and associated rubrics. Reviewers rated each task on five dimensions: question relevance, question scientific soundness, rubric reasoning depth, rubric scientific accuracy, and overall question quality. Only tasks rated ``Very Strong'' or ``Excellent'' for question relevance, question scientific soundness, rubric scientific accuracy, and overall quality were retained in the final dataset.

\begin{figure}[t]
\centering
\includegraphics[width=0.95\linewidth]{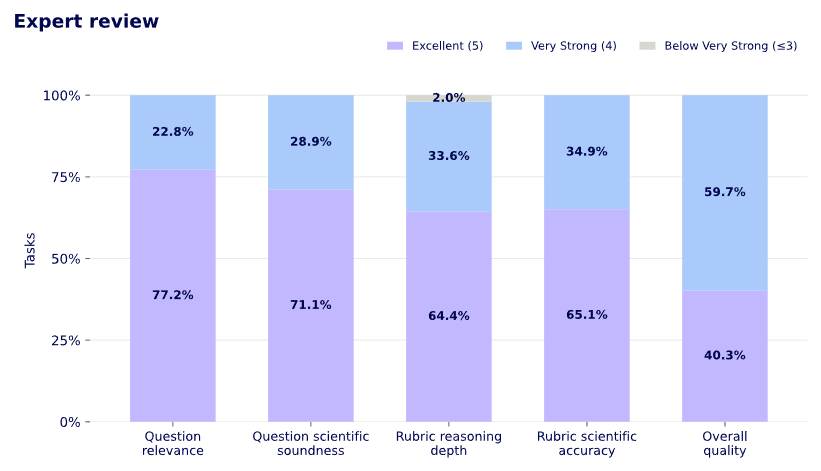}
\caption{Expert review ratings by assessment dimension. Share of retained tasks rated Excellent, Very Strong, or below Very Strong on each of the five dimensions.}
\label{fig:expert-review}
\end{figure}

Each task was reviewed by at least two independent reviewers, each of whom holds a PhD in a relevant field and has worked at least three years as a scientist in academia or industry. Figure~\ref{fig:expert-review} shows overall metrics across each measured dimension.

\begin{figure}[t]
\centering
\includegraphics[width=\linewidth]{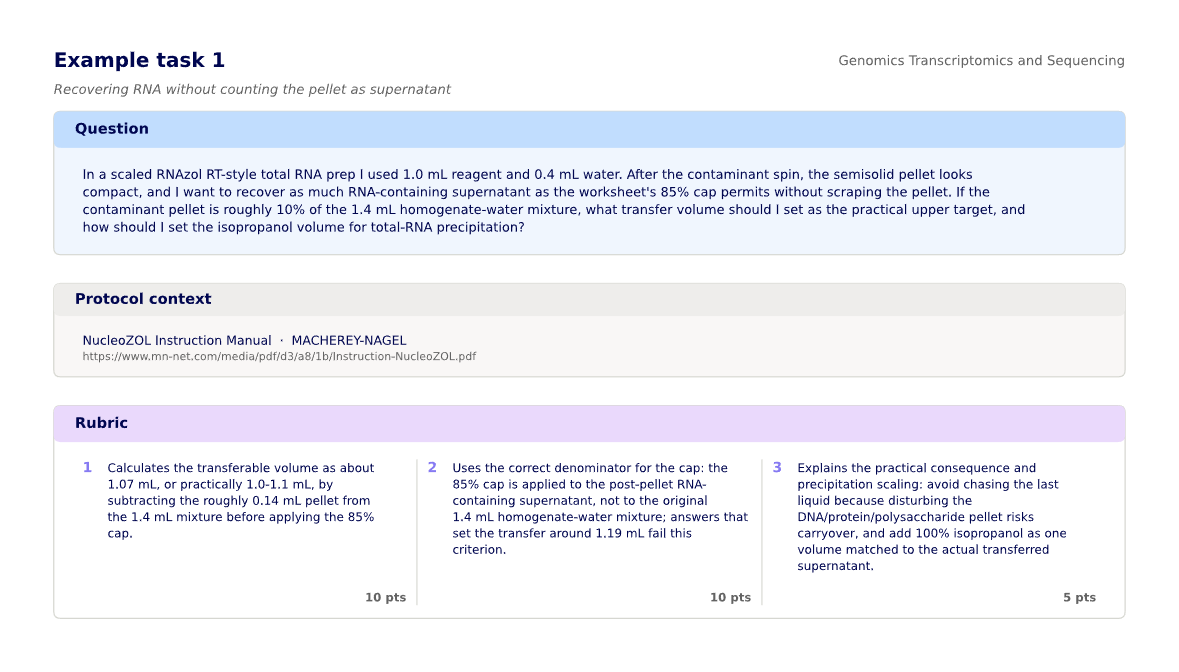}
\vspace{0.4em}
{\small \textbf{Example 1:} Sample task from genomics, transcriptomics, and sequencing. The scientist query, source protocol, and weighted rubric criteria are shown.}
\end{figure}

Additional sample tasks are shown in Appendix~\ref{app:sample-tasks}.

\section{Experimental Setup}
\label{sec:experimental-setup}

\subsection{Models Evaluated}

Nine models were evaluated: Claude Opus 5, GPT-5.6 Sol, GPT-5.6 Terra, GPT-5.6 Luna, Gemini 3.6 Flash, Grok 4.5, Kimi K3, GLM 5.2, and Inkling. All models were run at their highest available reasoning settings. Open models were run on Baseten, with exact settings detailed in Appendix~\ref{app:model-config}. Results are aggregated over ten attempts per task. Attempts that did not produce a valid response because of misformatted responses or refusals were assigned a score of zero; invalid-response rates by model are reported in Appendix~\ref{app:invalid-responses}.

\subsection{Evaluation Method}
\label{sec:evaluation-method}

Each model receives the query, a published protocol, and internet access, and returns a single free-text response as described in Section~\ref{sec:task-formulation}. Claude Opus, GPT-series, Gemini, and Grok models had access to their providers' first-party search tools, while open models were given internet access through Exa configured at ``auto'' effort. Tool use was unrestricted within those configurations. LLM graders were not given web search or tools. Appendix~\ref{app:grader-agreement} reports representative agreement results between two grader models.

\subsection{Metrics}

Model performance is reported using a normalized rubric score between 0 and 1. Each rubric criterion is scored as 0, 1, or 2, where 2 is full credit and 1 is partial credit. Each criterion additionally has a weight reflecting its relative importance to the overall task. A coarse three-point scale was used instead of a wider numeric range to improve LLM-as-a-judge stability. A response's normalized score is the sum of weighted criterion scores divided by the total weighted points available, and the average across ten attempts per task is reported. Confidence intervals are calculated by bootstrap resampling at the source-protocol level because tasks derived from the same protocol are not independent. The expected \bestatk{} score is computed by sampling $k$ of the ten attempts per question without replacement, taking the maximum normalized score, and averaging across resamples.

\section{Results}
\label{sec:results}

\begin{figure}[t]
\centering
\includegraphics[width=\linewidth]{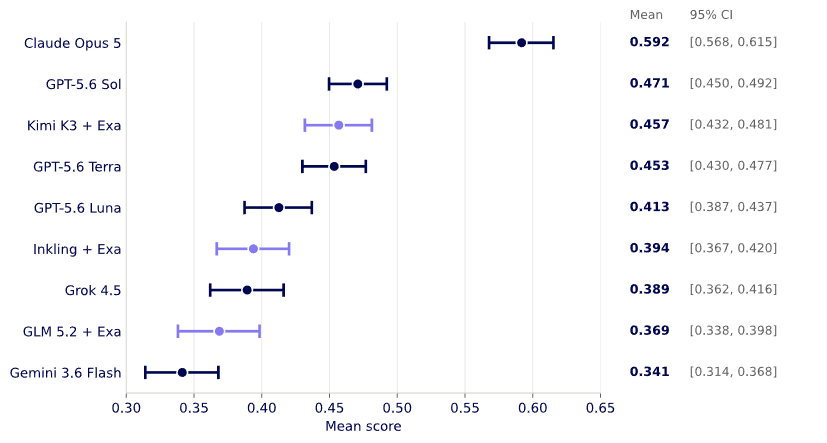}
\caption{Mean normalized rubric score by model. Means across 149 tasks are shown with 95\% confidence intervals.}
\label{fig:mean-score}
\end{figure}

Performance on BenchBench-Protocol varies substantially across models, task intents, and source-protocol domains. Claude Opus 5 is the strongest overall model, with a mean normalized score of 59.2\% (95\% CI: 56.8--61.5\%). Scores for the other evaluated models range from 34.1\% to 47.1\%. Scores for all evaluated models are shown in Figure~\ref{fig:mean-score}. BenchBench-Protocol has at least 40\% headroom of unearned weighted rubric credit across model families, allowing future frontier-model assessment.

\subsection{Performance Under Repeated Sampling}

Because a scientist can compare several candidate answers before acting on one, expected \bestatk{} performance is reported as an upper bound on single-model performance.

The ranking of model performance is not stable in $k$. GPT-5.6 Sol and Kimi K3 are statistically indistinguishable at $k=1$, but separate cleanly at $k=10$, with Kimi K3 performing better. Because absolute gains are bounded by the unearned credit at $k=1$, gains are reported as a fraction of headroom captured. Complete results are in Appendix~\ref{app:performance}; Appendix~\ref{app:score-distributions} shows the distribution of model performance across tasks.

\begin{figure}[t]
\centering
\includegraphics[width=0.9\linewidth]{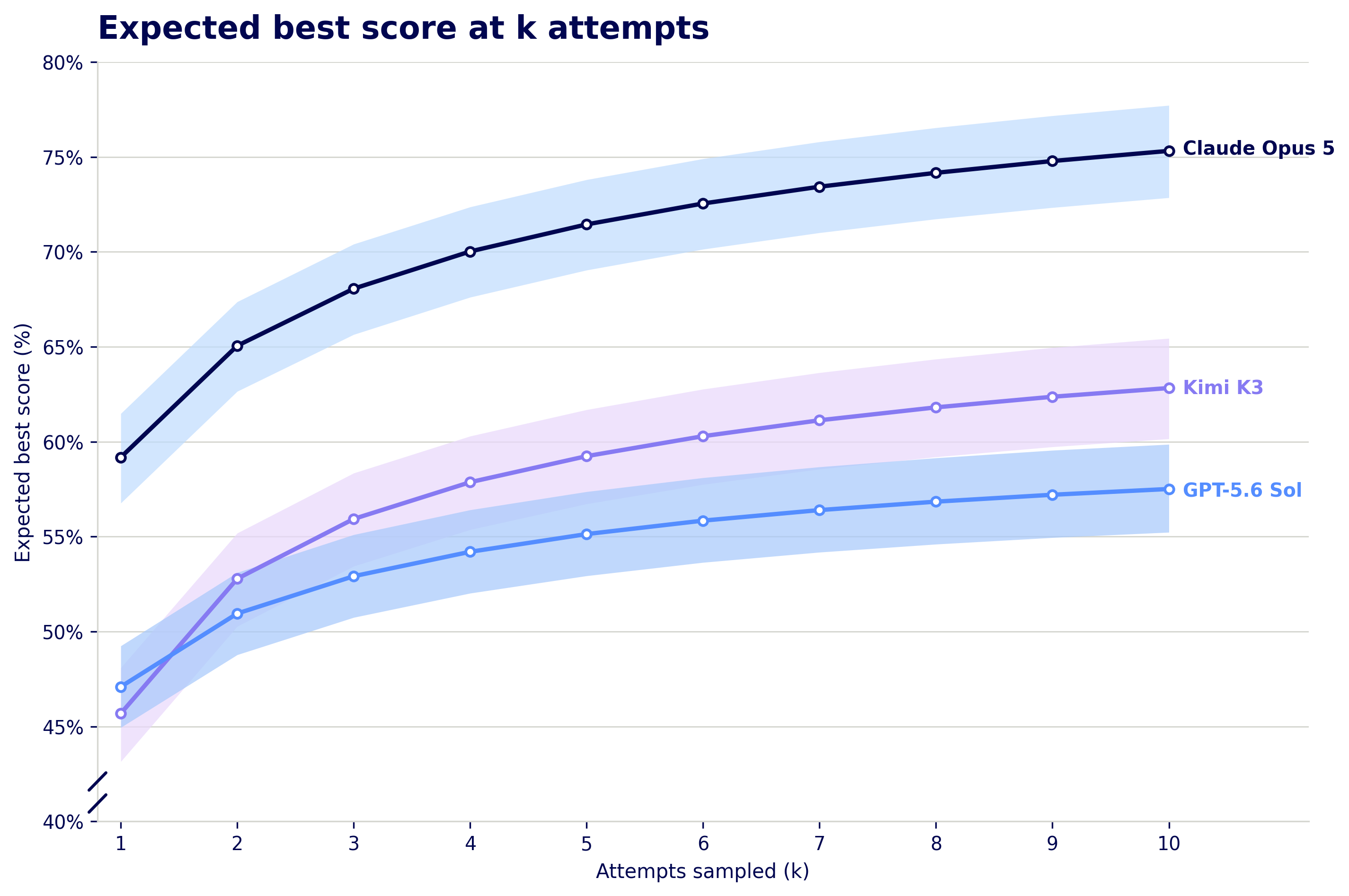}
\caption{Expected \bestatk{} score for three representative models. Expected maximum score is shown for $k$ of ten attempts per task with 95\% confidence bands. Complete nine-model results are in Table~\ref{tab:best-of-k}.}
\label{fig:best-of-k}
\end{figure}

At ten attempts, Claude Opus 5 captures 39.5\% of available headroom and Kimi K3 captures 31.5\%, while all three GPT-5.6 models capture between 19.6\% and 22.5\%. Models that capture less headroom over multiple attempts tend to be more predictable. Under oracle selection, the strongest model still leaves roughly one quarter of rubric credit unearned, indicating that the benchmark is not saturated at any $k$.

\subsection{Performance by Task Intent}

\begin{figure}[t]
\centering
\includegraphics[width=\linewidth]{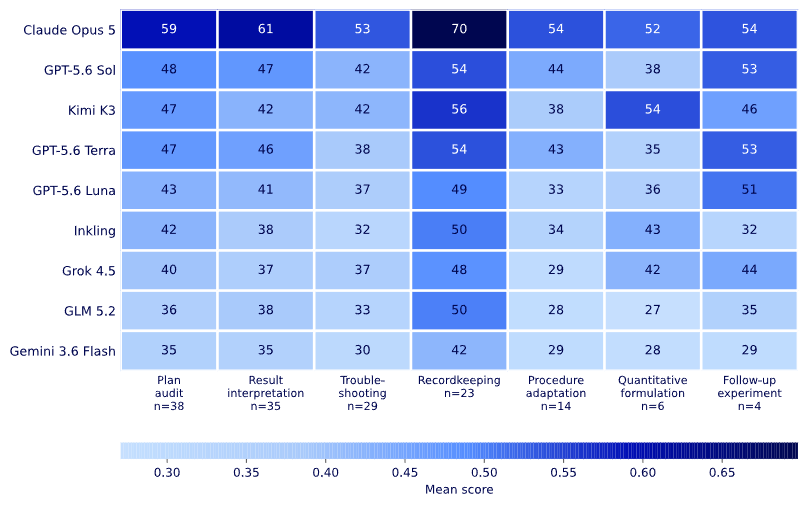}
\caption{Mean normalized rubric score by primary task intent. Cells give mean scores in percent, with task counts per intent.}
\label{fig:intent-performance}
\end{figure}

Figure~\ref{fig:intent-performance} demonstrates strong differences in performance across task intents.

Analyzing failure cases heuristically, models perform comparatively well on tasks that require reasoning from experimental evidence to the conclusions it supports. Models perform comparatively worse on tasks that require tacit knowledge or intuition about what an expert would consider the ``most likely'' or ``best'' approach, which may account for the discrepancy in troubleshooting and adaptation.

\subsection{Performance by Source-Protocol Domain}

\begin{figure}[t]
\centering
\includegraphics[width=\linewidth]{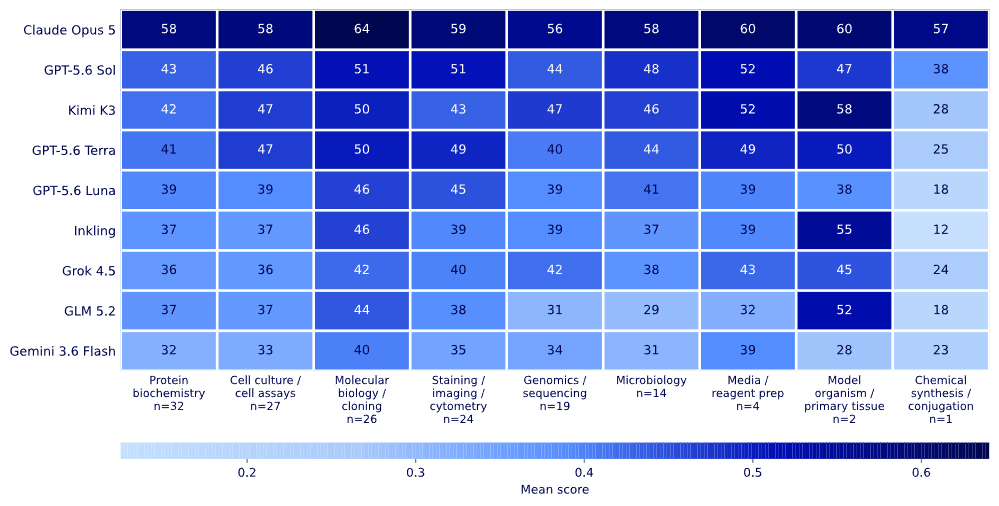}
\caption{Mean normalized rubric score by source-protocol domain. Cells give mean scores in percent, with task counts per domain.}
\label{fig:domain-performance}
\end{figure}

Claude Opus and GPT-series models perform strongly on staining and imaging tasks relative to open-source models, which may be due to general improvements in multimodal reasoning in these models versus open series. Figure~\ref{fig:domain-performance} summarizes these results; full values are reported in Table~\ref{tab:domain-performance}.

\subsection{Model Token Usage and Cost}

\begin{figure}[t]
\centering
\includegraphics[width=\linewidth]{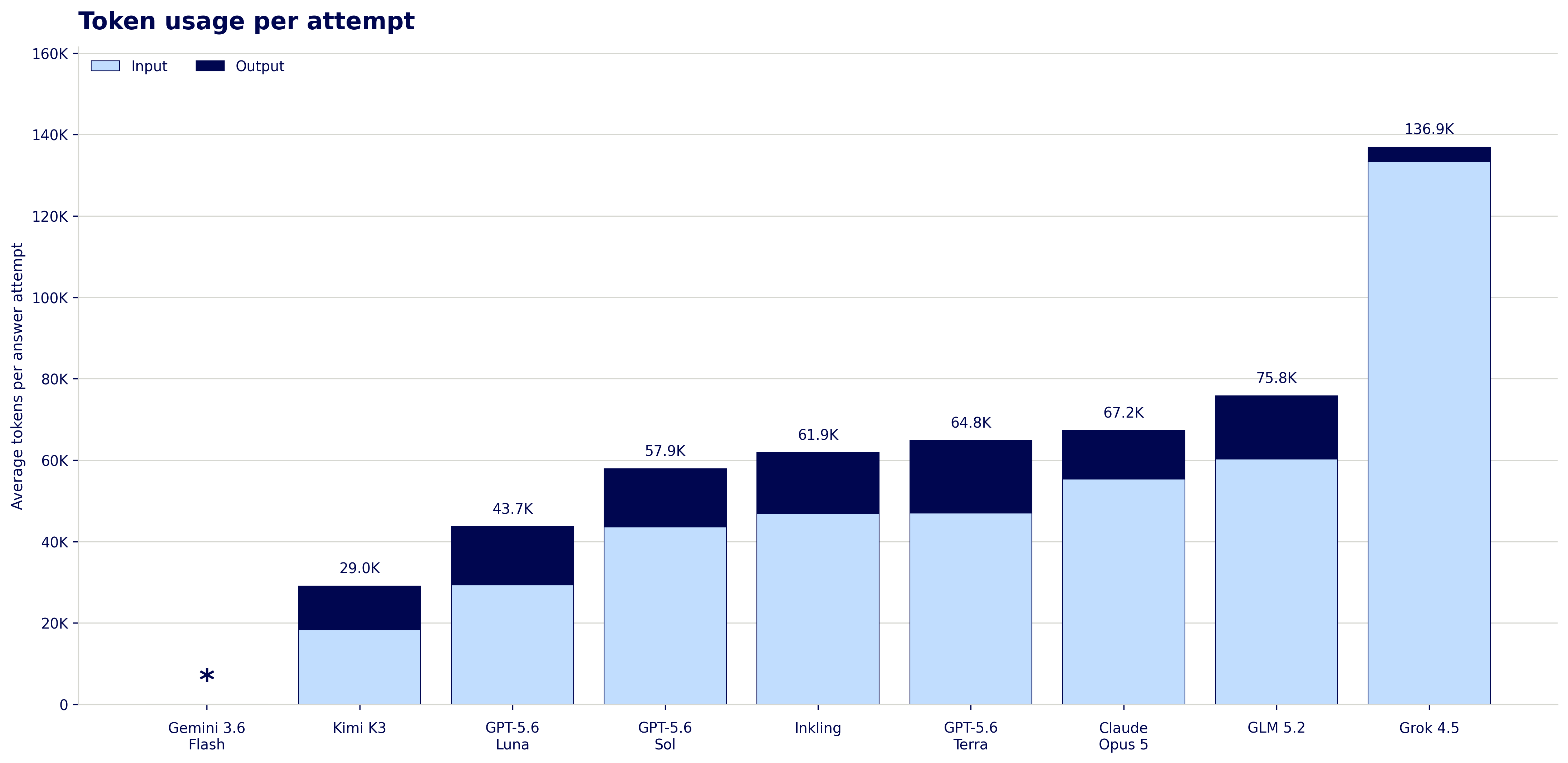}
\caption{Token usage per answer attempt. Average input and output tokens are shown per model; Gemini 3.6 Flash is excluded (asterisk).}
\label{fig:token-usage}
\end{figure}

Figure~\ref{fig:token-usage} visualizes average input and output token consumption across the benchmark task suite. Grok 4.5 consumes the most input tokens, likely because of a high volume of web searches and ingested content. At maximum reasoning settings, models perform frequent searches, leading to high volumes of input and output tokens. Gemini 3.6 Flash is excluded because search input tokens from Gemini search are billed separately and not precisely delineated in Vertex API calls, making total-input-token comparisons unfair. Average cost per answer attempt across model families, excluding web search and hosted-tool charges, is included in Appendix~\ref{app:cost}. The performance--cost Pareto frontier at maximum reasoning is included in Appendix~\ref{app:pareto}.

\section{Limitations and Future Work}

\subsection{Limitations}

BenchBench-Protocol measures reasoning about realistic modifications to wet-lab protocols, but it does not directly measure whether models improve experimental outcomes at the bench. The benchmark contains 149 tasks across nine domains of wet-lab biology, and while scientists perform the bulk of their protocol work in these areas, no single benchmark can span the full range of modifications that arise in this setting. Tasks were also selected to be difficult, so scores may not reflect performance on routine adjustments that make up the majority of day-to-day laboratory work.

The evaluation was conducted in a single-turn setting, with the model's output as the final response. Real use is almost always multi-turn, with scientists clarifying limitations in reagents, instruments, and time. Model performance may improve if the model is allowed to ask clarifying questions.

Model evaluation used the maximum reasoning settings offered by each model family to standardize the greatest available performance. This results in substantially higher token usage and cost than default or recommended settings, which affects the performance--cost frontier. Models may also perform differently in other harnesses or multi-agent setups.

Human expert baselines were not directly assessed on the benchmark tasks. The expert quality-control process places strong emphasis on answerability, which is a proxy adopted in place of a direct expert-performance baseline.

Responses were graded by GPT-5.6 Terra, which is also one of the nine evaluated models. Model grading of open-ended responses is imperfect in ways a coarse rubric scale reduces but does not remove, and the three-point scale itself limits how finely a response can be distinguished from a stronger one. Search tooling also differs between conditions, with closed models using first-party search and open models using Exa at ``auto'' effort.

\subsection{Future Work}

Future work can focus further on multi-turn settings that mirror how a wet-lab scientist may collaborate with a model. These can include tasks that rely on previous data or experiments to plan follow-up work, assess models' abilities to troubleshoot or debug real experimental failures, or correctly elicit user intent when underspecified. Further analysis of the existing benchmark can be conducted to understand failure modes of current models and assess performance in consensus or multi-model setups.

\section{Conclusion}

In this paper, we make the following contributions:
\begin{itemize}
    \item \textbf{Reviewed benchmark task corpus:} We introduce BenchBench-Protocol, 149 protocol-modification tasks spanning nine domains of wet-lab biology, recovered from modifications scientists made to published protocols during real experimental work. Tasks and rubrics were independently assessed by multiple scientists holding a PhD in a relevant domain and three or more years of bench experience.
    \item \textbf{Rubrics:} We provide weighted rubrics grounded in the modification that was actually performed, scoring responses against the specific downstream dependencies a correct answer must address rather than against general plausibility.
    \item \textbf{Model benchmarking:} We evaluate nine models across all 149 tasks, under repeated sampling as well as single attempts, and characterize their failures.
\end{itemize}

BenchBench-Protocol measures whether models can adapt an existing protocol while tracking the downstream steps a change affects. Claude Opus 5 scores highest at 59.2\%, with other models' scores ranging from 34.1\% to 47.1\%, and repeated sampling captures between one fifth and two fifths of unearned credit.

Kimi K3 approaches and then exceeds GPT-5.6 Sol in expected \scoreatk{} at ten attempts. Heuristic analysis of traces and responses shows that GPT-5.6 Sol had more consistent responses and more closely followed exact protocol constraints, while Kimi K3 had greater variation, producing both weaker and more complete responses. The heavier upper tail of scores gave Kimi K3 a higher expected \bestatk{} score.

Models differ in overall level rather than in profile, so current systems are not differentially suited to particular areas of wet-lab reasoning. Protocol modification is routine work where errors propagate into downstream results, and grounding evaluation in modifications scientists actually made offers a more direct way to measure progress on it.

\subsubsection*{Acknowledgments}

We thank Lauren DeVos for her support. We thank expert scientist contributors for providing real-world protocols and validating tasks and rubrics. Their work and contributions were invaluable for creating BenchBench-Protocol.

\bibliography{references}
\bibliographystyle{iclr2026_conference}

\appendix
\makeatletter
\@addtoreset{figure}{section}
\@addtoreset{table}{section}
\makeatother
\renewcommand{\thefigure}{\Alph{section}\arabic{figure}}
\renewcommand{\thetable}{\Alph{section}\arabic{table}}

\section{Protocol Domain and Task Intent Definitions}
\label{app:taxonomy}

\subsection{Protocol Domains}

Each source protocol is assigned exactly one domain. Where a protocol spans two domains, we assign the domain corresponding to the step in which the task's modification falls.

\begin{description}
\item[Protein biochemistry and purification:] Expression, extraction, chromatographic or affinity purification, buffer exchange, concentration, and biophysical or biochemical characterization of proteins. Includes enzymatic assays where the analyte is a purified protein. Excludes protein detection performed as a readout of a cell-based experiment, which is assigned to cell culture and cell-based assays.
\item[Cell culture and cell-based assays:] Maintenance, passaging, transfection, transduction, differentiation, and treatment of cultured cells, and assays whose readout is a property of the cell population. Includes viability, proliferation, reporter, and cytotoxicity assays. Excludes primary tissue and animal work, which is assigned to model organism, in vivo, and primary tissue.
\item[Molecular biology and cloning:] Nucleic acid manipulation: PCR, restriction and ligation, assembly cloning, mutagenesis, plasmid preparation, transformation, and sequence verification of constructs. Excludes library preparation intended for high-throughput sequencing, which is assigned to genomics, transcriptomics, and sequencing.
\item[Staining, imaging, and cytometry:] Sample fixation, permeabilization, labeling with antibodies or dyes, mounting, microscopy, image acquisition, flow cytometry, and analysis of the resulting images or cytometry data.
\item[Genomics, transcriptomics, and sequencing:] Nucleic acid extraction for sequencing, library preparation, quantification and quality control of libraries, sequencing runs, and associated sample handling. Excludes downstream computational analysis performed independently of sample handling.
\item[Microbiology and microbial methods:] Culture, isolation, enumeration, and characterization of bacteria, yeast, or other microorganisms, including growth curves, plating and colony counting, competent cell preparation, and antimicrobial susceptibility testing.
\item[Media, reagent prep, and recipe:] Preparation of media, buffers, stock solutions, and reagent mixtures, including formulation, sterilization, storage, and quality control, where the protocol's product is the reagent rather than an experimental result.
\item[Model organism, in vivo, and primary tissue:] Procedures on live animals or organisms, and harvest, dissociation, or handling of primary tissue, including dosing, husbandry-adjacent steps required by the protocol, and tissue processing before downstream assay.
\item[Chemical synthesis and conjugation:] Synthesis, modification, or conjugation of small molecules and bioconjugates, including labeling chemistry, purification of synthetic products, and characterization by chemical rather than biological means.
\end{description}

\subsection{Task Intents}

Each task is additionally assigned a single primary intent, defined as the intent corresponding to the rubric criteria carrying the largest share of total weight.

\begin{description}
\item[Plan audit:] The task requires evaluating a proposed or existing experimental plan for correctness, completeness, or internal consistency, and identifying what is missing, misordered, or unsupported.
\item[Measurement and result interpretation:] The task requires interpreting a quantitative or qualitative result (e.g., a yield, curve, gel, plot, or control outcome) and stating what it supports or rules out.
\item[Troubleshooting:] The task presents a failed, degraded, or anomalous outcome and requires identifying probable causes and the corrective action or diagnostic step that discriminates among them.
\item[Recordkeeping:] The task requires specifying what must be documented, versioned, or tracked for a modification to be interpretable or reproducible later, including lot numbers, instrument settings, deviations, and sample provenance.
\item[Procedure specification and adaptation:] The task requires modifying or extending a procedure for a new goal, sample type, scale, or reagent, and specifying resulting steps concretely enough to execute.
\item[Quantitative formulation:] The task requires a calculation where the numerical answer and correctness of its derivation are the object of assessment.
\item[Follow-up experiment:] The task requires proposing a subsequent experiment, including what it would measure and what its outcome would establish.
\end{description}

\subsection{Labeling Procedure}

Labels were assigned by GPT-5.6 Sol at extra-high reasoning. Task classifications were reviewed to ensure consistency and accuracy.

\section{Model Inference Configurations}
\label{app:model-config}

Closed systems used provider-hosted APIs; Kimi K3, GLM 5.2, and Inkling were served through Baseten. All systems used the highest reasoning tier exposed by their provider.

\begin{table}[h]
\caption{Model inference configurations. Access route, reasoning tier, and search provider are listed per model.}
\label{tab:model-config}
\centering
\scriptsize
\begin{tabularx}{\linewidth}{lYll}
\toprule
\textbf{Model} & \textbf{Access} & \textbf{Reasoning} & \textbf{Search provider} \\
\midrule
Claude Opus 5 & Anthropic API & max & First party / native \\
GPT-5.6 Sol & OpenAI API & max & First party / native \\
GPT-5.6 Terra & OpenAI API & max & First party / native \\
GPT-5.6 Luna & OpenAI API & max & First party / native \\
Gemini 3.6 Flash & Vertex AI & HIGH & First party / native \\
Grok 4.5 & xAI Responses & high & First party / native \\
Kimi K3 & Baseten (\texttt{moonshotai/Kimi-K3}) & max & Exa ``auto'' \\
GLM 5.2 & Baseten (\texttt{zai-org/GLM-5.2}) & max & Exa ``auto'' \\
Inkling & Baseten (\texttt{thinkingmachines/inkling}) & max & Exa ``auto'' \\
\bottomrule
\end{tabularx}
\end{table}

\section{Alternate Grader Agreement}
\label{app:grader-agreement}

We regraded 20 tasks with Claude Opus 5 under an identical system prompt. Exact agreement was 81.2\%, and within-one-level agreement was 99.95\%, with Claude Opus 5 being a slightly more permissive grader.

\begin{table}[h]
\caption{Grader agreement on a 20-task subset. Mean scores under each grader are shown with paired shift and 95\% confidence interval.}
\label{tab:grader-agreement}
\centering
\scriptsize
\begin{adjustbox}{max width=\linewidth}
\begin{tabular}{lrrrr}
\toprule
\textbf{Model} & \textbf{Terra} & \textbf{Opus} & \textbf{Shift} & \textbf{95\% CI} \\
\midrule
Claude Opus 5 & 0.550 & 0.656 & +0.106 & [0.084, 0.130] \\
GPT-5.6 Sol & 0.476 & 0.549 & +0.073 & [0.046, 0.104] \\
Kimi K3 & 0.432 & 0.498 & +0.066 & [0.034, 0.101] \\
GPT-5.6 Terra & 0.465 & 0.513 & +0.048 & [0.022, 0.076] \\
GPT-5.6 Luna & 0.425 & 0.479 & +0.054 & [0.031, 0.081] \\
Inkling & 0.383 & 0.442 & +0.059 & [0.032, 0.088] \\
Grok 4.5 & 0.391 & 0.432 & +0.041 & [0.012, 0.074] \\
GLM 5.2 & 0.305 & 0.330 & +0.025 & [0.006, 0.045] \\
Gemini 3.6 Flash & 0.295 & 0.338 & +0.043 & [0.024, 0.065] \\
\midrule
Pooled across systems & 0.414 & 0.471 & +0.057 & [0.038, 0.078] \\
\bottomrule
\end{tabular}
\end{adjustbox}
\end{table}

\clearpage
\section{Distribution of Average Performance by Model}
\label{app:score-distributions}

\begin{figure}[h]
\centering
\includegraphics[width=\linewidth]{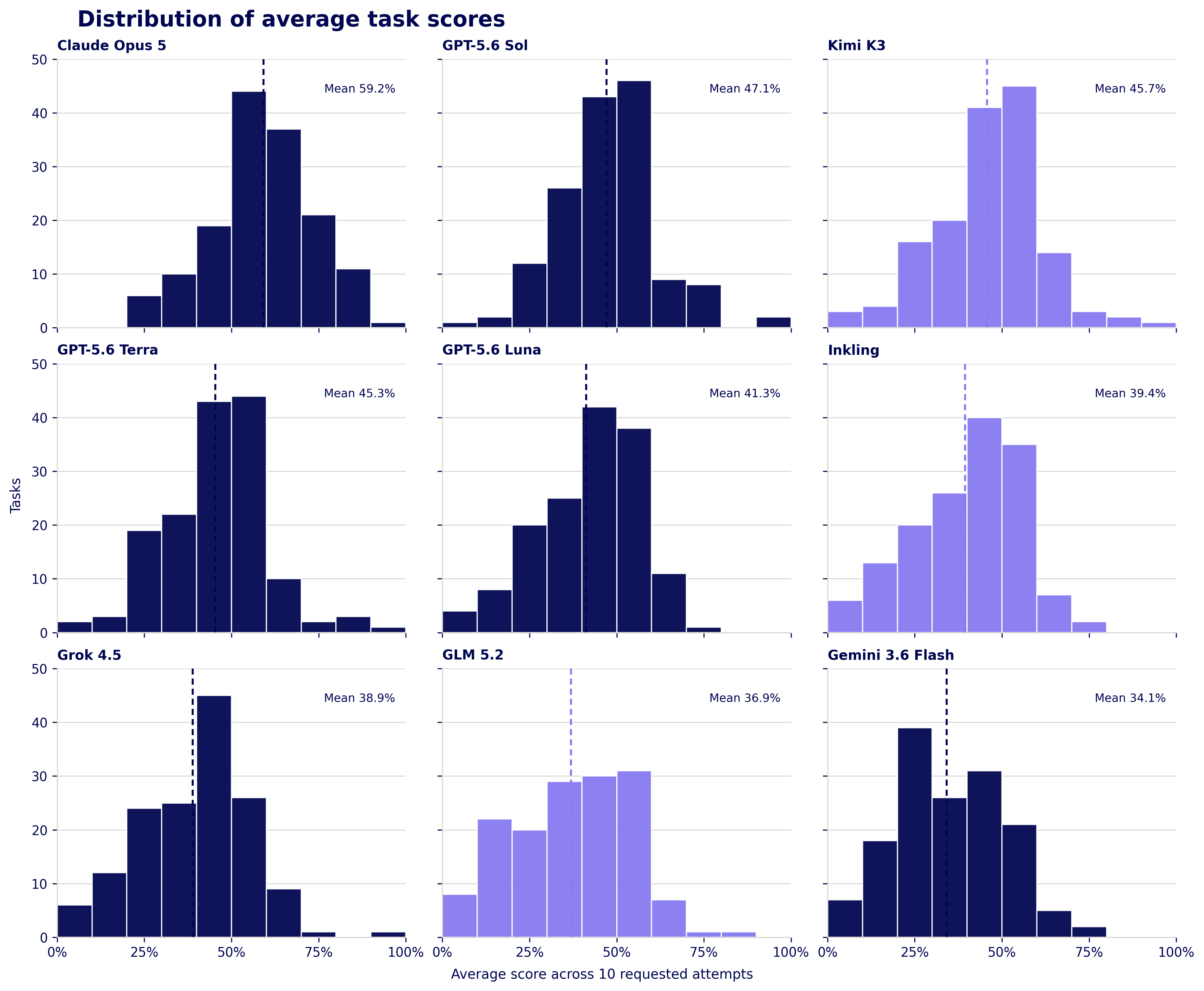}
\caption{Distribution of average task scores by model. Per-task means across ten attempts are shown for each model, with the model mean marked.}
\label{fig:score-distributions}
\end{figure}

\clearpage
\begin{landscape}
\section{Performance by Source-Protocol Domain and Expected \texorpdfstring{\bestatk{}}{best-of-k} Metrics}
\label{app:performance}

\begin{table}[h]
\caption{Performance by source-protocol domain. Mean normalized rubric score is shown per domain with 95\% confidence intervals.}
\label{tab:domain-performance}
\centering
\scriptsize
\begin{adjustbox}{max width=\linewidth}
\begin{tabular}{lccccccccc}
\toprule
\textbf{Model} & \textbf{Protein} & \textbf{Cell} & \textbf{Molecular} & \textbf{Staining} & \textbf{Genomics} & \textbf{Microbiology} & \textbf{Media} & \textbf{Organism} & \textbf{Chemical} \\
\midrule
Claude Opus 5 & .581 [.526,.634] & .584 [.531,.638] & .640 [.580,.692] & .589 [.534,.644] & .561 [.489,.626] & .583 [.535,.641] & .603 [.448,.758] & .596 & .567 \\
GPT-5.6 Sol & .430 [.382,.471] & .465 [.417,.515] & .510 [.452,.578] & .508 [.469,.552] & .438 [.371,.497] & .479 [.402,.540] & .515 [.508,.523] & .471 & .375 \\
Kimi K3 & .416 [.361,.461] & .467 [.390,.537] & .496 [.446,.541] & .430 [.374,.489] & .470 [.407,.536] & .463 [.378,.545] & .520 [.500,.540] & .582 & .275 \\
GPT-5.6 Terra & .413 [.360,.459] & .466 [.403,.532] & .501 [.451,.566] & .487 [.438,.533] & .405 [.338,.469] & .439 [.372,.502] & .492 [.463,.521] & .500 & .250 \\
GPT-5.6 Luna & .394 [.339,.444] & .387 [.316,.455] & .461 [.400,.514] & .448 [.403,.490] & .387 [.316,.455] & .414 [.335,.479] & .395 [.394,.396] & .382 & .183 \\
Inkling & .374 [.326,.417] & .367 [.298,.434] & .461 [.401,.515] & .394 [.330,.457] & .388 [.317,.457] & .371 [.249,.471] & .393 [.360,.425] & .550 & .125 \\
Grok 4.5 & .363 [.299,.420] & .365 [.291,.435] & .422 [.350,.498] & .397 [.338,.454] & .417 [.356,.476] & .376 [.299,.462] & .427 [.404,.450] & .446 & .242 \\
GLM 5.2 & .372 [.305,.433] & .368 [.285,.449] & .441 [.367,.502] & .383 [.318,.446] & .306 [.259,.355] & .293 [.197,.393] & .322 [.256,.387] & .525 & .183 \\
Gemini 3.6 Flash & .318 [.257,.375] & .328 [.243,.411] & .399 [.341,.450] & .346 [.298,.390] & .342 [.277,.402] & .310 [.215,.400] & .387 [.344,.431] & .279 & .233 \\
\bottomrule
\end{tabular}
\end{adjustbox}
\end{table}
\end{landscape}
\clearpage

\begin{table}[h]
\caption{Expected \bestatk{} metrics. Scores at $k=1,2,5,$ and $10$ are shown with the gain from $k=1$ to $k=10$.}
\label{tab:best-of-k}
\centering
\small
\begin{tabular}{lrrrrrr}
\toprule
\textbf{System} & \textbf{@1} & \textbf{@2} & \textbf{@5} & \textbf{@10} & \textbf{Gain} & \textbf{95\% CI @10} \\
\midrule
Claude Opus 5 & .592 & .651 & .714 & .753 & +.161 & [.728,.777] \\
GPT-5.6 Sol & .471 & .509 & .551 & .575 & +.104 & [.552,.598] \\
Kimi K3 & .457 & .528 & .592 & .628 & +.171 & [.601,.654] \\
GPT-5.6 Terra & .453 & .497 & .546 & .577 & +.123 & [.550,.603] \\
GPT-5.6 Luna & .413 & .453 & .498 & .527 & +.115 & [.501,.553] \\
Inkling & .394 & .449 & .506 & .541 & +.147 & [.512,.569] \\
Grok 4.5 & .389 & .445 & .499 & .533 & +.144 & [.499,.565] \\
GLM 5.2 & .369 & .435 & .503 & .549 & +.181 & [.521,.577] \\
Gemini 3.6 Flash & .341 & .394 & .446 & .476 & +.134 & [.448,.503] \\
\bottomrule
\end{tabular}
\end{table}

\clearpage
\section{Additional Sample Tasks}
\label{app:sample-tasks}

\begin{figure}[h]
\centering
\includegraphics[width=\linewidth]{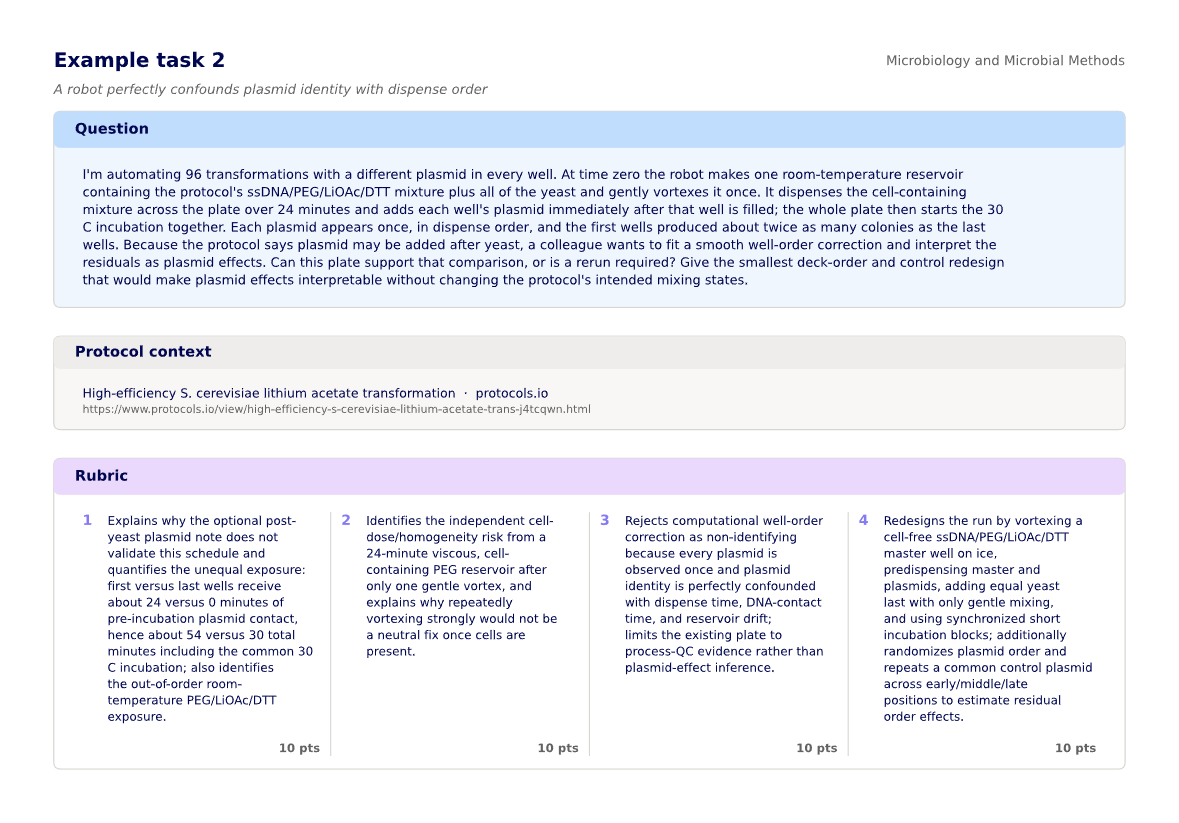}
\vspace{0.4em}
{\small \textbf{Example F1:} Sample task from microbiology and microbial methods. The scientist query, source protocol, and weighted rubric criteria are shown.}
\end{figure}

\begin{figure}[h]
\centering
\includegraphics[width=\linewidth]{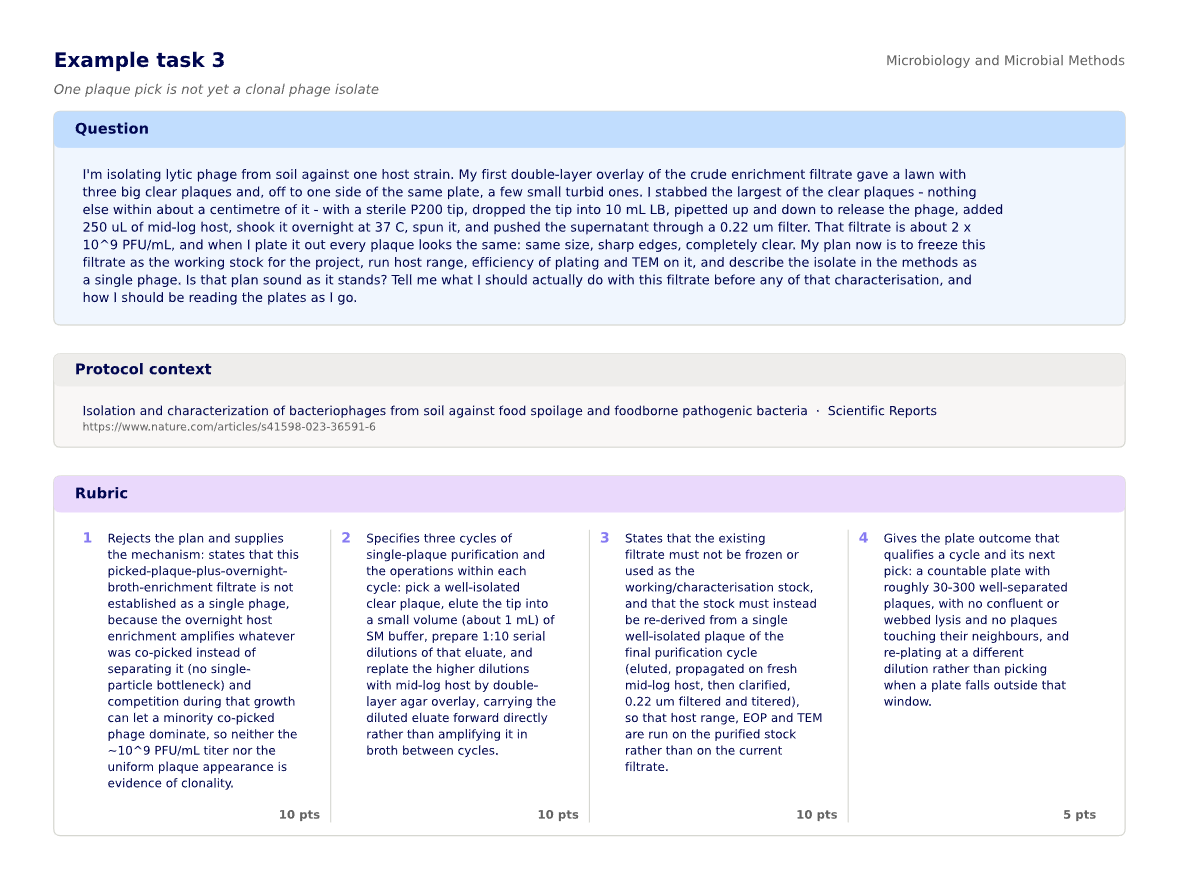}
\vspace{0.4em}
{\small \textbf{Example F2:} Sample task from microbiology and microbial methods. The scientist query, source protocol, and weighted rubric criteria are shown.}
\end{figure}

\begin{figure}[h]
\centering
\includegraphics[width=\linewidth]{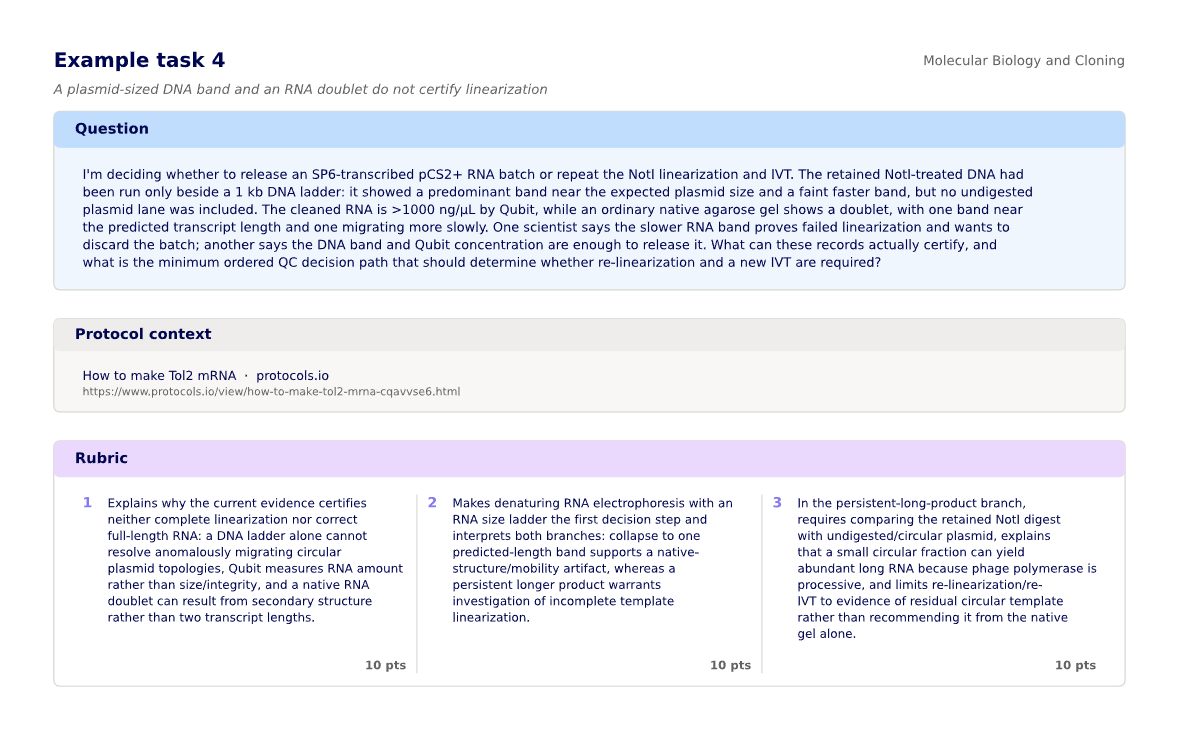}
\vspace{0.4em}
{\small \textbf{Example F3:} Sample task from molecular biology and cloning. The scientist query, source protocol, and weighted rubric criteria are shown.}
\end{figure}

\begin{figure}[h]
\centering
\includegraphics[width=\linewidth]{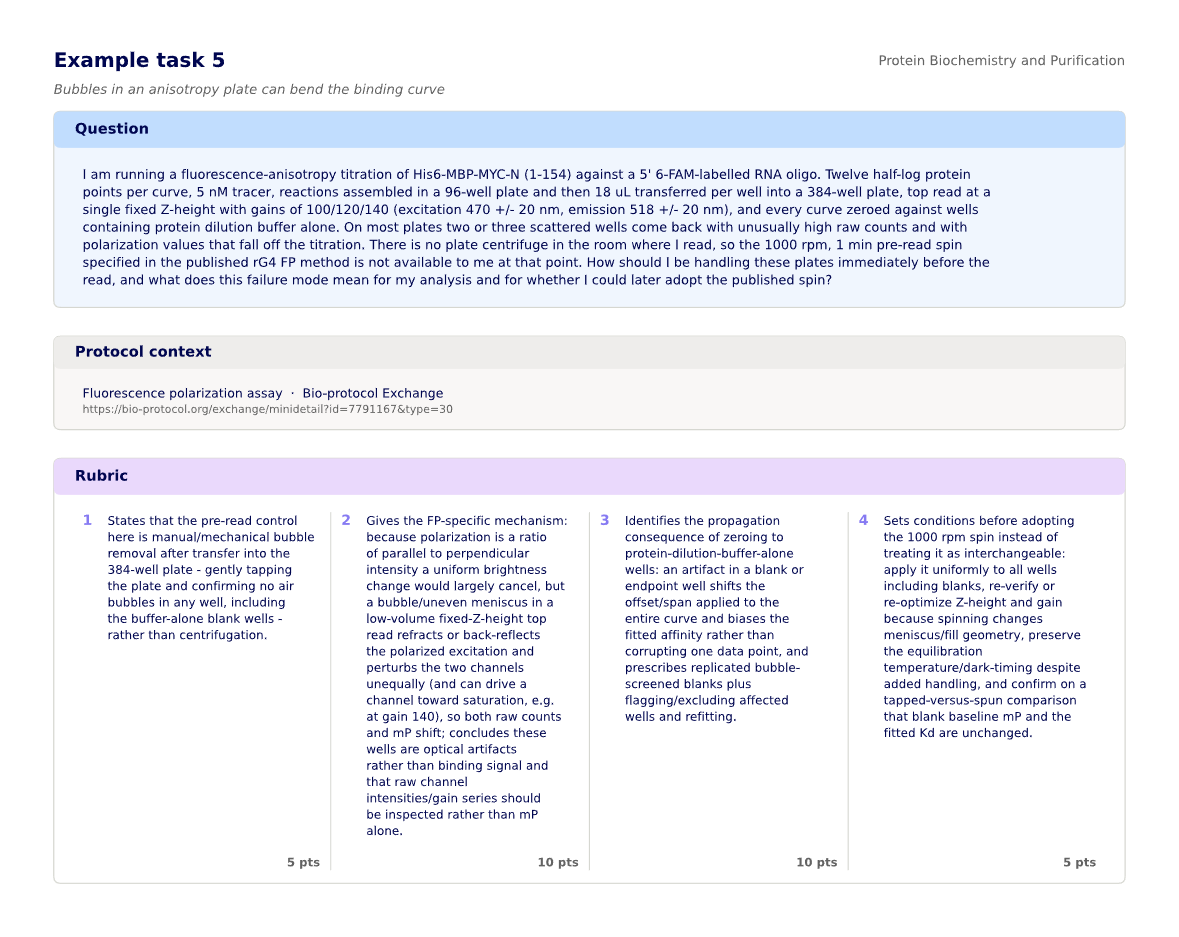}
\vspace{0.4em}
{\small \textbf{Example F4:} Sample task from protein biochemistry and purification. The scientist query, source protocol, and weighted rubric criteria are shown.}
\end{figure}

\clearpage
\section{Average Answer-Generation Cost per Attempt}
\label{app:cost}

\begin{figure}[h]
\centering
\includegraphics[width=\linewidth]{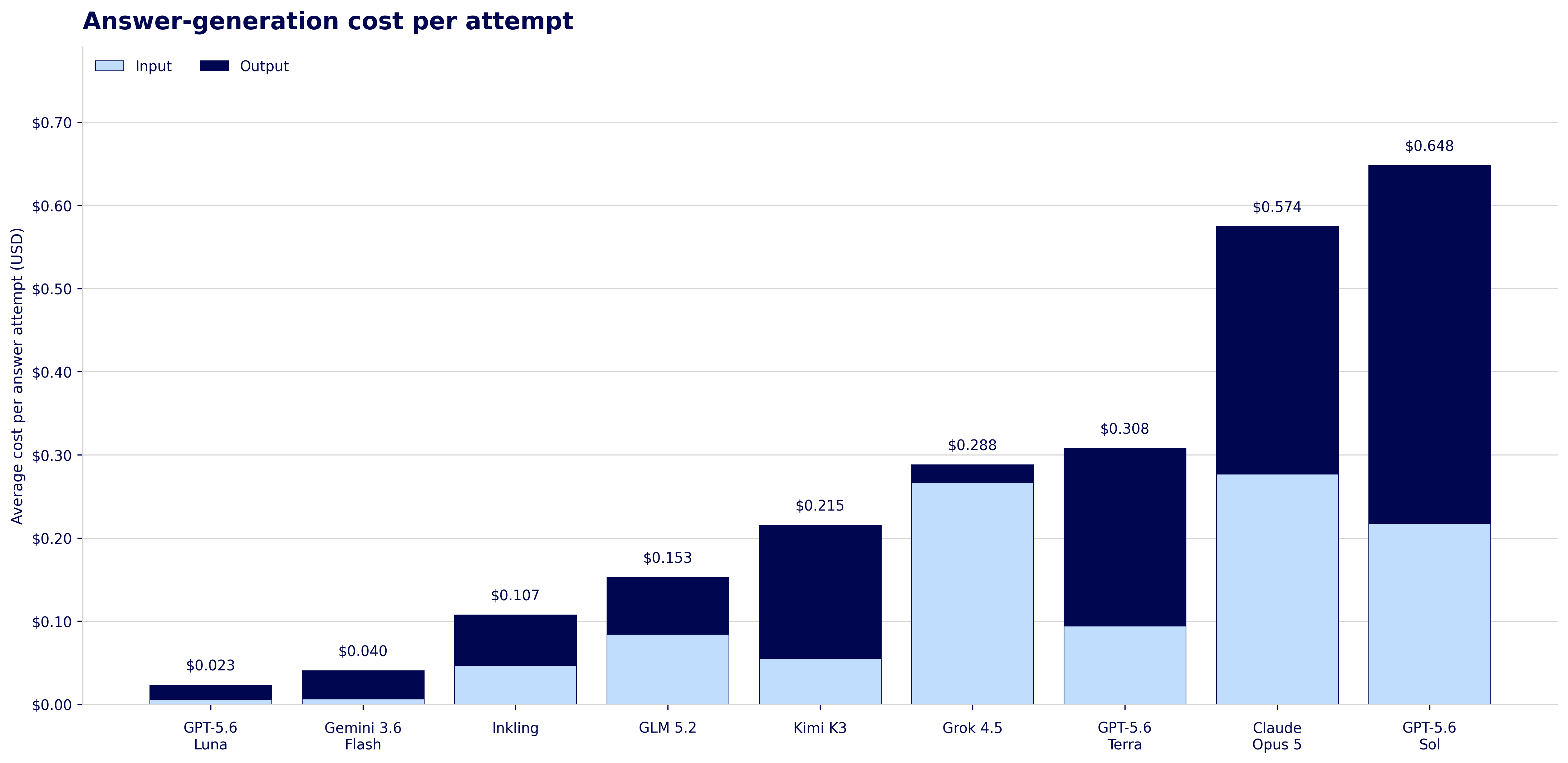}
\caption{Answer-generation cost per attempt. Average input and output cost is shown per model, excluding web search and hosted-tool usage.}
\label{fig:cost-per-attempt}
\end{figure}

Average cost per attempt across model families was computed using input and output token counts and API pricing on August 13, 2026, for Gemini 3.6 Flash, GPT-series models, Claude Opus 5, and Grok 4.5. Pricing for GLM 5.2, Inkling, and Kimi K3 is representative pricing from Baseten's model inference API service, collected on August 13, 2026. Costs exclude web search or hosted-tool usage.

\clearpage
\section{Performance--Cost Frontier}
\label{app:pareto}

\begin{figure}[h]
\centering
\includegraphics[width=\linewidth]{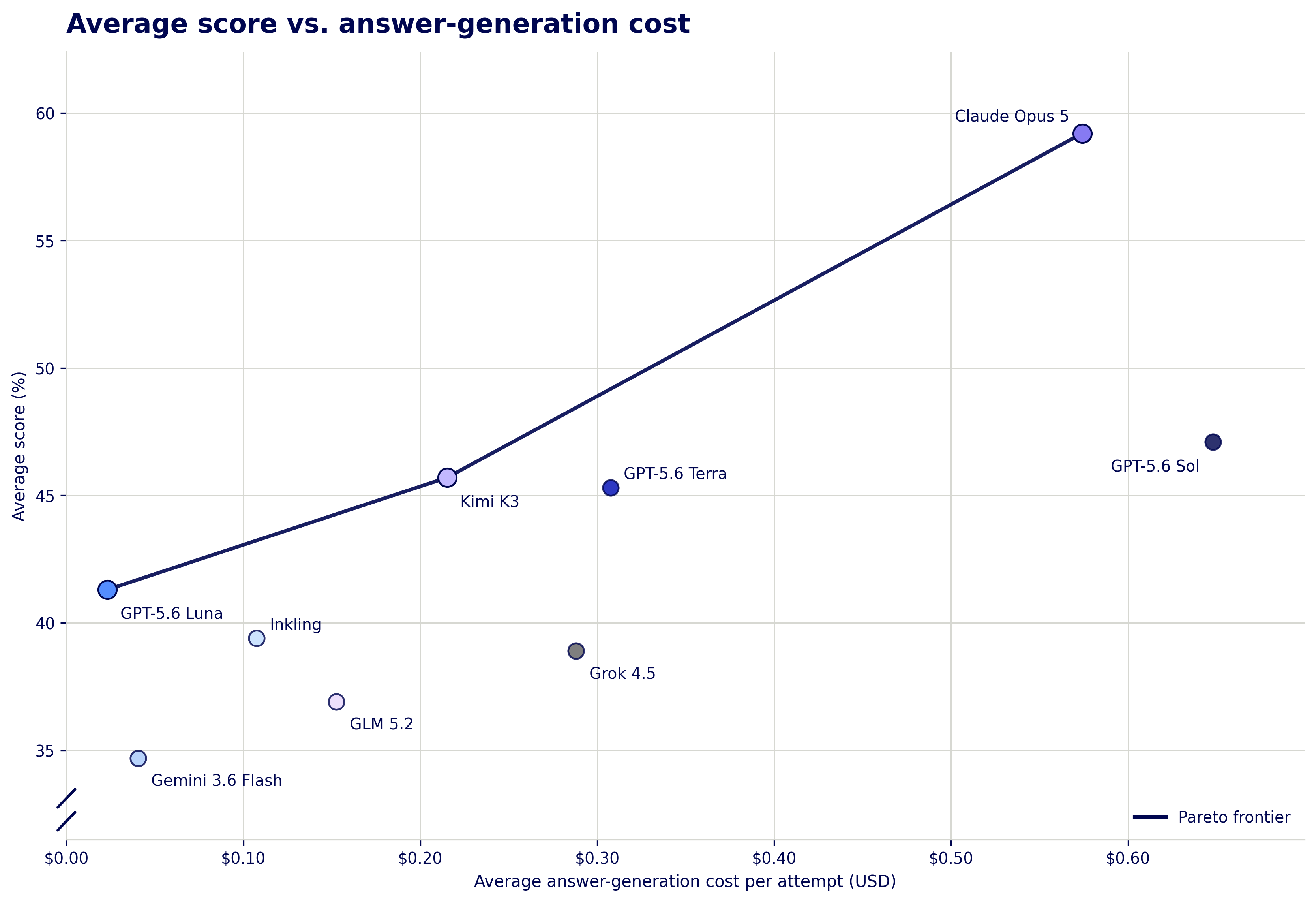}
\caption{Average score against answer-generation cost per attempt. The Pareto frontier is drawn over GPT-5.6 Luna, Kimi K3, and Claude Opus 5; the score axis is truncated.}
\label{fig:pareto}
\end{figure}

Figure~\ref{fig:pareto} shows that GPT-5.6 Luna, Kimi K3, and Claude Opus 5 lie on the performance--cost Pareto frontier. Claude Opus 5 achieves the highest score (59.2\%) while costing less than GPT-5.6 Sol, and Kimi K3 slightly outperforms GPT-5.6 Terra at lower cost. All models were assessed at their maximum reasoning settings, which can strongly affect costs and therefore the Pareto frontier.

\clearpage
\section{Model Misformatted Responses and Refusals}
\label{app:invalid-responses}

\begin{table}[h]
\caption{Rate of refusals and misformatted outputs across model families.}
\label{tab:invalid-responses}
\centering
\small
\begin{tabular}{lrrr}
\toprule
\textbf{Model} & \textbf{Misformatted} & \textbf{Refusals} & \textbf{Total} \\
\midrule
Claude Opus 5 & 0 / 1,490 & 0 / 1,490 & 0.0\% \\
GPT-5.6 Sol & 0 / 1,490 & 0 / 1,490 & 0.0\% \\
Kimi K3 & 62 / 1,490 & 0 / 1,490 & 4.2\% \\
GPT-5.6 Terra & 0 / 1,490 & 0 / 1,490 & 0.0\% \\
GPT-5.6 Luna & 0 / 1,490 & 0 / 1,490 & 0.0\% \\
Inkling & 28 / 1,490 & 0 / 1,490 & 1.9\% \\
Grok 4.5 & 10 / 1,490 & 0 / 1,490 & 0.7\% \\
GLM 5.2 & 181 / 1,490 & 0 / 1,490 & 12.1\% \\
Gemini 3.6 Flash & 0 / 1,490 & 106 / 1,490 & 7.1\% \\
\bottomrule
\end{tabular}
\end{table}

GLM 5.2 produced misformatted responses on 181 of 1,490 attempts (12.1\%), followed by Kimi K3 on 62 attempts (4.2\%), Inkling on 28 attempts (1.9\%), and Grok 4.5 on 10 attempts (0.7\%). Gemini 3.6 Flash refused 106 of 1,490 requests (7.1\%). Claude Opus 5 and all three GPT-5.6 models produced valid responses for every attempt. Unsuccessful attempts because of refusals or misformatting were assigned a score of zero.

\end{document}